\documentclass{webofc}

\usepackage[varg]{txfonts}   
\usepackage{hyperref}
\usepackage{url}
\usepackage{amsmath}
\usepackage{tikz}
\usetikzlibrary{patterns, arrows.meta}
\usetikzlibrary{decorations.pathreplacing}
\usepackage{graphicx}
\usepackage{caption}

\hypersetup{colorlinks=true,citecolor=blue,urlcolor=blue,linkcolor=blue}

\begin{document}
\title{Energy-aware frugal Bayesian optimization} 
%
%

\author{\firstname{Gaston} \lastname{Plat}\inst{1,2,3}\fnsep\thanks{Corresponding author e-mail: 
{\href{mailto:gaston.plat@onera.fr}{gaston.plat@onera.fr}}}
\and
        \firstname{Paul} \lastname{Saves}\inst{4}\fnsep
        \and
        \firstname{Nathalie} \lastname{Bartoli}\inst{1,3}\fnsep
        \and 
        \firstname{Thierry} \lastname{Lefebvre}\inst{1,3}\fnsep
        \and
        \firstname{Joseph} \lastname{Morlier}\inst{2, 3}\fnsep
}

\institute{DTIS, ONERA, Université de Toulouse, Toulouse, 31000, France.
\and
           Université de Toulouse, ISAE-SUPAERO Toulouse, 31000, France 
\and 
           Fédération ENAC ISAE-SUPAERO ONERA, Université de Toulouse, Toulouse, 31000, France. 
\and
        IRIT, UMR 5505 CNRS, Université Toulouse Capitole, Toulouse, 31000, France
          }

\abstract{
Modern design optimization frameworks aim first and foremost for models with the most accurate predictions without balancing computational overhead. It remains a reason why scaled architecture and multidisciplinary design optimization problems are difficult to address, even with sample-efficient Bayesian optimizers. In this paper, a metric quantifying the computational energy footprint is introduced within a Bayesian optimization framework to guide the parameter setting of a model towards configurations that balance both performance and frugality. The computer experiments highlighted existing tradeoffs between optimum convergence and the underlying energy footprint, and sometimes resulted in both a better-found optimum and lower energy consumption. 
}

%
\maketitle
\section{Introduction}
\label{intro}
In engineering design, frugality within models is often traded for better predictive capabilities during the preliminary design stage. Furthermore, the design process often relies on costly and derivative-free computer experiments that are used to explore innovative concepts. Given a set of inputs, the models opaquely output engineering metrics, such as mechanical stresses in an aircraft, and are treated as expensive-to-evaluate black-box functions~\cite{audet2017derivative}. In this context, architecture optimization problems, whose design variables involve inter-dependencies~\cite{saves2025modelinghierarchicalspacesreview} and different input data types~\cite{paul_mdo_mixed}, often lead to intractable computations because the optimum search must hierarchically explore heterogeneous high-dimensional design spaces~\cite{bussemaker2020system}, and address complex coupled subproblems with intricate variables~\cite{halle2025distance}. The computational burden is therefore an inherent challenge in engineering design. 

One of the critical challenges in frugal optimization for engineering is to find the best optimal designs requiring the least computational resources. Especially in Bayesian optimization (BO)~\cite{frazier2018bayesian}, any optimization process is executed with a given budget, which is often the number of black-box evaluations. Therefore, frugal optimization frameworks must navigate through the current informativeness of the dataset~\cite{frazier2018bayesian}, the informativeness of the next sequence of sampled points~\cite{lee_nonmyopic_2021}, the prediction quality of the models~\cite{forrester2007multi}, the costs of the models~\cite{charayron2023towards}, the computational machinery~\cite{hota2018survey}, the grid supply capabilities~\cite{hanifi_advanced_2024}, and the remaining computational budget to determine what is the most resource-effective convergence trajectory towards the true engineering optimum. 

As a result, the field lacks a formal framework to bridge optimal design with efficient computational platform~\cite{lannelongue2021green, candelieri2024fair}. This paper addresses this gap by introducing a novel metric based on the energy footprint of computations during optimization processes. As hardware and operating systems (OS) become more advanced, open-source features and libraries are now able to measure the instantaneous power consumption of hardware~\cite{weaver_measuring_2012}. These tools are integrated within a BO framework to develop a computationally energy-sensitive optimum search. This study aims to highlight how the automated, energy-aware selection of model parameters can save energy while maintaining accurate convergence. 

The main contributions of this paper are twofold: first, defining a metric to assess the electrical energy footprint of convergence trajectories; and second, identifying Pareto optimal parameter settings to manage the tradeoff between computational cost and optimization performance. 
We validate this framework first on a toy optimization problem, and then on a benchmark engineering application: minimizing the mass of a 10-bar truss~\cite{rostum2026comparative}. By embedding automated parameter selection into the BO framework, this study reduces the reliance on manual hyperparameter tuning and demonstrates how deeply parameter configurations dictate convergence behavior. Because this sensitivity is complex and lacks established empirical rules~\cite{feurer2019hyperparameter}, leveraging an automated method is essential for achieving both resource efficiency and high accuracy in engineering design.
%

The remainder of the paper is organized as follows. Section~\ref{methodo} describes the mathematical background of the method. Section~\ref{results} highlights the numerical results on the tradeoffs between frugality and performance of the convergence trajectories, and Section~\ref{discussions} concludes with a summary and directions for future research.

\section{Methodology}
\label{methodo}

\subsection{Initial design of experiments}
\label{OLHD}
This step aims at selecting the initial points with the most valuable information regarding the rest of the optimization process. To do so, Latin hypercube sampling (LHS) is leveraged to have a uniform yet frugal starting dataset for our computer experiments. More specifically, let $D \in \mathbb{N}^*$ denote the dimensionality of the problem at hand. The initial LHS  design of experiments (DoE) is obtained by sampling $n_0 \in \mathbb{N^*}$ initial points distributed across $n_0$ non-intersecting hypercubes, that consequently partition the design space into $n_0$ subspaces.  Choosing the hypercube locations is a space-filling problem, solved by optimizing the permutation of the columns of the initial design matrix $X_{DoE} \in \mathbb{R}^{n_0 \times D}$ according to a space-filling criterion~\cite{wu2019space}. In this study, the $\phi_p$ criterion is used with $p=10$ since it can be formulated analytically with ease of computation.
\begin{equation}
    \phi_p(X_{DoE}) = \left(\sum_{1 \leq i < j \leq n_0}d_{ij}^{-p}(X_{DoE})\right)^{1/p}, 
\end{equation}
where $d_{ij}(\cdot) \in \mathbb{R}$ is the intersited Euclidean distance between the $i$-th and $j$-th column of the initial design matrix $X_{DoE}$. The enhanced stochastic evolutionary (ESE) algorithm~\cite{jin2005efficient} is used to obtain a desirable DoE that satisfies the requirements of uniformity and subprojection~\cite{wu2019space}. 

\subsection{Gaussian processes for regression}
\label{GPs}
\subsubsection{Regular Gaussian processes}
\label{GP}
Unlike traditional deterministic predictors, such as linear regressors, Gaussian process (GP)~\cite{rasmussen_gaussian_2008} accounts for uncertainty and outputs a pointwise Gaussian distribution on the quantity of interest. Given a problem of dimensionality $D$, a dataset $\mathcal{D}=\{X, \textbf{y}\}$ of $n$ points with inputs $X \in \mathbb{R}^{n\times D}$ and targets  $\textbf{y}\in \mathbb{R}^n$, a GP defines a distribution over functions $f(\textbf{x})\sim \mathcal{GP}\ \bigl(   m(\textbf{x}), k(\textbf{x}, \textbf{x}') \bigr) $ fully described by a mean function $m(\cdot)$ and a covariance function $k(\cdot, \cdot)$. The GP takes advantage of the Bayesian inference scheme
where a posterior value is inferred from prior assumptions about relationships between variables in the design space. The covariance function, also called kernel, encodes this prior information about correlation between variables. In this framework, the squared exponential kernel $k$ is given by 
\begin{equation}
    k(\textbf{x}, \textbf{x}') = \sigma^2_f\prod_{i=1}^{D}\exp\left(-\frac{1}{2l_i^2}(x_i-x_i')^2\right),  
    \label{kernelGP}
\end{equation}
where $(\textbf{x},\textbf{x}') \in \left(\mathbb{R}^D\right)^2$, $(x_i,x_i') \in \mathbb{R}^2$ are the $i$-th components of the \textbf{x} and \textbf{x}' vectors, $\sigma_f$ is the signal variance and $l_i$ is the characteristic lengthscale in the $i$-th dimension. This kernel encodes assumptions on  smoothness and correlation lengthscales. The quality of GP predictions is then influenced by its set of output variance, and lengthscales $\Theta=\{\sigma_f, l_1, ..., l_D\}$. The optimal set is commonly chosen by maximizing a closed-form expression of the logarithmic marginal likelihood with respect to $\Theta$. The predictive distribution of $n_*$ number of unobserved design points $X_*\in \mathbb{R}^{n_* \times D}$ follows a multivariate Gaussian distribution with a mean and a covariance that are in closed-form too. 
Thus, GP regression is fast, tractable, and shows good data feature-extracting capabilities, especially for interpolation.
%

%

\subsubsection{GP handling hierarchical relationships between variables}
\label{hierGP}
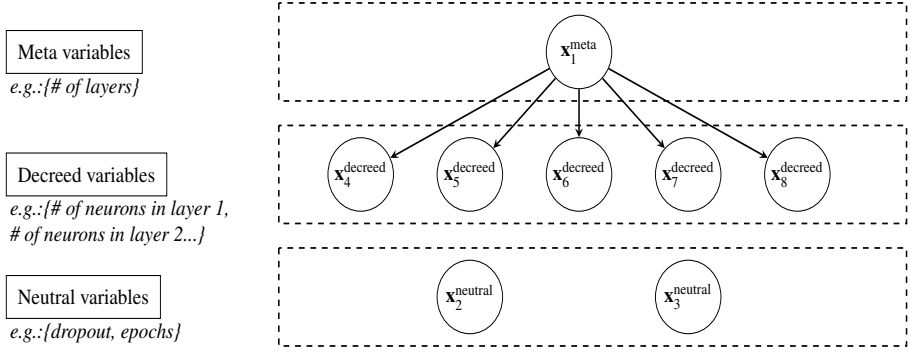
\begin{figure}[thb!]
    \centering
    \resizebox{0.92\columnwidth}{0.35\columnwidth}{%
    \begin{tikzpicture}[
        var/.style={circle, draw, minimum size=1.2cm, inner sep=0pt},
        labelbox/.style={rectangle, draw, minimum height=0.7cm, inner sep=6pt},
        arrow/.style={->, >=stealth, thick},
        dasharrow/.style={->, >=stealth, dashed, thick}
    ]

    \draw[dashed, thick] (-1.5, 0.8) rectangle (10, -0.8);    
    \draw[dashed, thick] (-1.5, -1.2) rectangle (10, -2.8);   
    \draw[dashed, thick] (-1.5, -3.2) rectangle (10, -4.8);   

    \node[var] (m1) at (4, 0) {$\textbf{x}_1^{\text{meta}}$};


    \node[var] (dec1) at (0, -2) {$\textbf{x}_4^{\text{decreed}}$};
    \node[var] (dec2) at (2, -2) {$\textbf{x}_5^{\text{decreed}}$};
    \node[var] (dec3) at (4, -2) {$\textbf{x}_6^{\text{decreed}}$};
    \node[var] (dec4) at (6, -2) {$\textbf{x}_7^{\text{decreed}}$};
    \node[var] (dec5) at (8, -2) {$\textbf{x}_8^{\text{decreed}}$};

    \node[var] (neu1) at (2, -4) {$\textbf{x}_2^{\text{neutral}}$};
    \node[var] (neu2) at (6, -4) {$\textbf{x}_3^{\text{neutral}}$};

    \node[labelbox, anchor=west] (l_m) at (-6.5, 0) {Meta variables};
    \node[anchor=north west, font=\itshape, inner sep=2pt] at (l_m.south west) {\textit{e.g.}:\{\# of layers\}};


    \node[labelbox, anchor=west] (l_dec) at (-6.5, -2) {Decreed variables};
    \node[anchor=north west, font=\itshape, inner sep=2pt, align=left] at (l_dec.south west) {\textit{e.g.}:\{\# of neurons in layer 1, \\ \# of neurons in layer 2...\}};

    \node[labelbox, anchor=west] (l_neu) at (-6.5, -4) {Neutral variables};
    \node[anchor=north west, font=\itshape, inner sep=2pt] at (l_neu.south west) {\textit{e.g.}:\{dropout, epochs\}};

    \draw[arrow] (m1) -- (dec1);
    \draw[arrow] (m1) -- (dec2);
    \draw[arrow] (m1) -- (dec3);
    \draw[arrow] (m1) -- (dec4);
    \draw[arrow] (m1) -- (dec5);


    \end{tikzpicture}
    
    }
    \caption{Taxonomy example of hierarchical inclusion-exclusion relationships. A basic example is given for the neural architecture of a multilayer perceptron (MLP).}    
    \label{fig:taxonomyhier}
\end{figure}

One can consider variables whose existence and range of values are conditioned on the state of another variable. It indicates a hierarchical relationship between the existence of variables. Consequently, they have roles that inform whether they are conditioned by a variable, conditioning others, or none~\cite{halle2025distance}, as in Figure~\ref{fig:taxonomyhier}. The role names are decreed, meta, or neutral, respectively.
This heterogeneous design space then changes the distance formulation between variables, and the GP kernel (see Eq.~\eqref{kernelGP}) needs to be adapted. In this study,  the distance $d(\cdot) \in \mathbb{R}$ used is introduced in~\cite{saves2025modelinghierarchicalspacesreview} as the algebraic distance:
\begin{equation}
    d(\textbf{x},\textbf{x}')=
    \begin{cases}
        1 & \text{if } \textbf{x}^T\textbf{x}'=0\\
        \frac{||\textbf{x}-\textbf{x}'||}{\sqrt{||\textbf{x}||^2+1}\sqrt{||\textbf{x}'||^2+1}} & \text{otherwise},
    \end{cases}
\end{equation}
where $||\textbf{x}-\textbf{x}'|| = \left(\sum_{i=1}^nd_i(x_i, x'_i)^2\right)^{1/2}$. $d_i(\cdot, \cdot) \in \mathbb{R}$ is the included-excluded distance~\cite{halle2025distance} taking into account hierarchical relationships between variables:
\begin{equation}
    d_i(x_i,x_i')=
    \begin{cases}
        x_i-x'_i & \text{if \textbf{x}, \textbf{x}' both included}\\
        0 & \text{if \textbf{x}, \textbf{x}' both excluded}\\
        \delta_i & \text{if only \textbf{x} or \textbf{x}' is excluded}.
    \end{cases}
    \label{hierdistance}
\end{equation}
Here, $\delta_i=\max\{x_i-x'_i\}/2 \geq 0$, as proven in~\cite{halle2025distance}. The kernel becomes symmetric positive definite~\cite{saves2025modelinghierarchicalspacesreview}, as required. Steps on selecting the hyperparameters $\Theta$ and building the predictive distribution remain unchanged.

\subsection{Unconstrained multi-objective Bayesian optimization 
%
}
\label{MOBO}
In the BO scheme, each objective is represented by a probabilistic surrogate model. In GP surrogate modeling, the $i$-th objective is modeled as $f_i(\textbf{x})\sim \mathcal{N}(\hat{\mu}_{f_i}(\textbf{x}), \hat{\sigma}_{f_i}(\textbf{x}))$, with $i \in [1, n_{obj}]$, design variable $\textbf{x} \in \mathcal{X}$, and design space $\mathcal{X} \subset \mathbb{R}^D$. Ultimately, the goal is to figure out which variable \textbf{x} minimizes all objectives as 
$\textbf{x}^* =  \text{arg}
    \min_{\textbf{x} \in \mathcal{X}} \left( f_1(\textbf{x}), f_2(\textbf{x}),..., f_{n_{obj}}(\textbf{x}) \right).
$
In this study, all multiobjective optima are found using Pareto optimality as define hereinafter~\cite{grapin2022regularized}.
\begin{itemize}
    \item  $\textbf{x}^{(1)}$ \textit{dominates} $\textbf{x}^{(2)}$ if it performs at least equally in every objective, that is $\forall i \in [1, n_{obj}], f_i(\textbf{x}^{(1)}) \leq f_i(\textbf{x}^{(2)})$, and if it exists at least one objective where it performs better: $\exists j \in [1, n_{obj}], f_j(\textbf{x}^{(1)}) < f_j(\textbf{x}^{(2)})$.  A non-dominated point is said to be \textit{Pareto optimal}.
    \item The \textit{Pareto set} (PS) is the collection of all Pareto optimal points. The \textit{Pareto front} (PF) is the image of the PS, that is, the collection of all Pareto optimal objective values.
\end{itemize}
To identify which point has the highest potential to bring the new information, Bayesian optimization is based on acquisition functions $\alpha(\cdot) \in \mathbb{R}$. Therefore, at each iteration, the next point to query with respect to the input variables is given by 
\begin{equation}
        \textbf{x}_{n+1} = \arg\max_{\textbf{x} \in \mathcal{X}} (\alpha(\textbf{x})).
        \label{eq:acquisition}
    \end{equation}
Two of such functions are described afterwards. 

\label{PI}
First, for each candidate $\textbf{x}$, the probability of improvement $\text{PI}(\textbf{x})$ is defined as the probability over the entire Pareto set that the point \textbf{x} is not dominated:
\begin{equation}
    \forall \textbf{u} \in \text{PS},  \quad \text{PI}(\textbf{x}) =  \mathbb{P}(\textbf{x} \preceq \textbf{u}).
    \label{eq:PIformulation}
\end{equation}
This criterion can be expressed in closed-form, following~\cite{charayron2023towards}:
\begin{equation}
    \mathbb{P}(\textbf{x}\preceq \textbf{u}) = 1 - \prod_{i=1}^{n_{obj}} \Phi \left(\frac{\hat{\mu}_{f_i}(\textbf{x})-f_i(\textbf{u})}{\hat{\sigma}_{f_i}(\textbf{x})}\right),
    \label{eq:closedformPI}
\end{equation}
where $\Phi$ is the cumulative distribution function of the standard normal distribution.
 
\label{EHVI}
Second, drawing inspiration from the single objective case, the hypervolume (HV)~\cite{guerreiro2021hypervolume} improvement is a generalization of the improvement formulation that leverages the Lebesgue measure $\Lambda$ to compare all $n_{obj}$ objectives at once. 
\begin{equation}
    \text{HVI}(\textbf{x}) = \max\left(0, \text{HV}(\{\textbf{x}, \text{PF} \})-\text{HV}(\text{PF})\right),
    \label{eq:HVI}
\end{equation}
\begin{equation}
    \text{HV}(\text{PF})=\Lambda\left(\bigcup_{\textbf{p} \in \text{PF}} [\textbf{p}, \textbf{R}]\right)
    \label{eq:HV},
\end{equation}
where $[\textbf{p}, \textbf{R}] = \{\textbf{q} \in \mathbb{R}^{n_{obj}}, \textbf{p} \prec \textbf{q} \quad \text{and} \quad \textbf{q} \prec \textbf{R}\}$ corresponds to the box delimited below by the Pareto front and above by the reference point $\textbf{R}$.
The expected hypervolume improvement (EHVI) can be used as an acquisition function
\begin{equation}
    \text{EHVI}(\textbf{x}) = \mathbb{E}[\text{HVI}(\textbf{x})] = \int_{-\infty}^{\infty} \text{HVI}(\textbf{x})\varphi(z)dz,
    \label{eq:EHVI}
\end{equation}
where $\varphi(\cdot)$ is the probability density of the normal distribution $\mathcal{N}(0,1)$. This formulation is no longer in a closed-form and implicitly assumes independence between the objective models. 

\section{Numerical results}
\label{results}

\subsection{Optimization setup}
\label{setup}
A multi-objective Bayesian optimizer with regular GP surrogates was used to identify Pareto-optimal hyperparameter configurations. The software used is the super-efficient global optimization with mixture of experts (SEGOMOE)~\cite{bartoli2019adaptive}. 
%
%
The logarithmic marginal likelihood of the GP surrogate models is maximized using the COBYLA~\cite{powell1998direct} algorithm with randomly generated initial candidates.
%
%
%
The acquisition function maximization stage is solved using the COBYLA algorithm with initial multi-start points sampled from an initial LHS DoE. 

Concerning the computational energy footprint measurement, the Intel running average power limit measures the energy consumed over a given time interval by an Intel or AMD central processing unit and its dynamic random access memory. So does the Nvidia management library for an Nvidia graphical processing unit. CodeCarbon~\cite{benoit_courty_2024_11171501}, a tool that unifies these libraries within a Python framework, is used to combine most of the hardware consumptions, and measurements are assumed to be exact. The computations were executed on a CPU-oriented computing architecture, using 32 GB DDR5 4800MHz of RAM. The reference of the CPU is "Sapphire Rapids" 8468, 48 cores, 2.1 GHz. 

\subsection{Toy model: parameter configuration for an accurate MLP classifier}
\label{mnist}
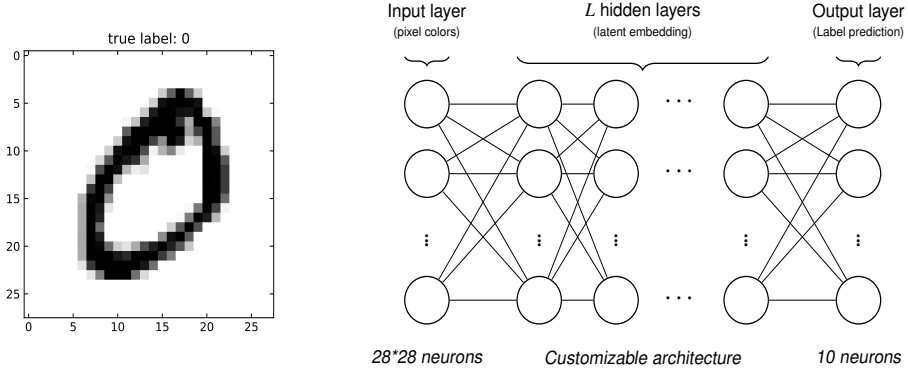
\begin{figure}[thbp]
    \centering
    \resizebox{0.95\columnwidth}{0.22\columnwidth}{%
    \begin{minipage}{0.24\textwidth} 
        \centering
        \includegraphics[width=\linewidth,clip]{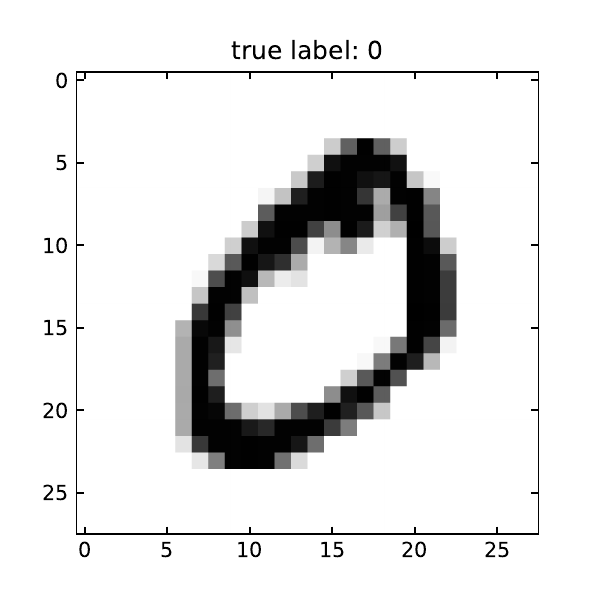}
        \label{digit}
    \end{minipage}%
    \hspace{0.5cm} 
    \begin{minipage}{0.40\textwidth} 
        \centering
        \raisebox{0.4cm}{%
        \resizebox{\linewidth}{!}{%
        \begin{tikzpicture}[
            neuron/.style={circle, draw, very thick, minimum size=1.5cm, inner sep=0pt, fill=white},
            dots/.style={font=\Huge},
            >=stealth
        ]
        
        \def\yA{3.2}
        \def\yB{1.0}
        \def\yDots{-1.0}
        \def\yC{-3.0}
        
        \def\xI{0}       
        \def\xHone{3.8}  
        \def\xHtwo{6.4}  
        \def\xDots{8.6}  
        \def\xHL{10.8}   
        \def\xO{14.6}    
        
        
        \node[neuron] (I1) at (\xI, \yA) {};
        \node[neuron] (I2) at (\xI, \yB) {};
        \node[dots] at (\xI, \yDots) {$\vdots$};
        \node[neuron] (I3) at (\xI, \yC) {};
        
        \node[neuron] (H11) at (\xHone, \yA) {};
        \node[neuron] (H12) at (\xHone, \yB) {};
        \node[dots] at (\xHone, \yDots) {$\vdots$};
        \node[neuron] (H13) at (\xHone, \yC) {};
        
        \node[neuron] (H21) at (\xHtwo, \yA) {};
        \node[neuron] (H22) at (\xHtwo, \yB) {};
        \node[dots] at (\xHtwo, \yDots) {$\vdots$};
        \node[neuron] (H23) at (\xHtwo, \yC) {};
        
        \node[dots] at (\xDots, \yA) {$\cdots$};
        \node[dots] at (\xDots, \yB) {$\cdots$};
        \node[dots] at (\xDots, \yC) {$\cdots$};
        
        \node[neuron] (HL1) at (\xHL, \yA) {};
        \node[neuron] (HL2) at (\xHL, \yB) {};
        \node[dots] at (\xHL, \yDots) {$\vdots$};
        \node[neuron] (HL3) at (\xHL, \yC) {};
        
        \node[neuron] (O1) at (\xO, \yA) {};
        \node[neuron] (O2) at (\xO, \yB) {};
        \node[dots] at (\xO, \yDots) {$\vdots$};
        \node[neuron] (O3) at (\xO, \yC) {};

        
        \foreach \i in {1,2,3} {
            \foreach \j in {1,2,3} {
                \draw[thick] (I\i) -- (H1\j);
            }
        }
        
        \foreach \i in {1,2,3} {
            \foreach \j in {1,2,3} {
                \draw[thick] (H1\i) -- (H2\j);
            }
        }
        
        \foreach \i in {1,2,3} {
            \foreach \j in {1,2,3} {
                \draw[thick] (HL\i) -- (O\j);
            }
        }

        
        \draw[decorate,decoration={brace,amplitude=8pt},very thick]
            (\xI+0.75, \yA+1.5) -- (\xI-0.75, \yA+1.5)
            node[midway, above=12pt, align=center, font=\sffamily\LARGE] {Input layer \\[-0.5ex] \large (pixel colors)};
        
        \draw[decorate,decoration={brace,amplitude=12pt},very thick]
            (\xHL+0.75, \yA+1.5) -- (\xHone-0.75, \yA+1.5)
            node[midway, above=12pt, align=center, font=\sffamily\LARGE] {$L$ hidden layers \\[-0.5ex] \large (latent embedding)};
        
        \draw[decorate,decoration={brace,amplitude=8pt},very thick]
            (\xO+0.75, \yA+1.5) -- (\xO-0.75, \yA+1.5)
            node[midway, above=12pt, align=center, font=\sffamily\LARGE] {Output layer \\[-0.5ex] \large (Label prediction)};

        
        \node[font=\sffamily\LARGE\itshape] at (\xI, \yC-1.8) {28*28 neurons};
        \node[font=\sffamily\LARGE\itshape] at ({(\xHone+\xHL)/2}, \yC-1.8) {Customizable architecture};
        \node[font=\sffamily\LARGE\itshape] at (\xO, \yC-1.8) {10 neurons};
        \end{tikzpicture}%
        }
        }
    \end{minipage}
    }
    \caption{MLP for 28x28 pixel digit image (MNIST~\cite{lecun2010mnist}) classification. 6000 (10\%) images are used for training, whereas 54000 are used for testing (90\%).} \label{fig:mlp} 

\end{figure}

In this study, the MLP classifier propagates digit images through latent and output layers to finally predict a label, as illustrated in Figure~\ref{fig:mlp}. During the perceptron training phase, the model weights are optimized with Adam~\cite{kingma2015adam}. The final predictive capabilities are highly sensitive to the number of iteration epochs, the number of layers and neurons, and the dropout ratio that prevents overfitting. These parameter values, commonly called hyperparameters~\cite{feurer2019hyperparameter}, are often highly coupled and require experts or automated methods to identify the best configurations; the latter process is referred to as hyperparameter optimization (HPO). Although previous studies have already tried to reduce training time~\cite{smithson2016neural} with respect to parameter settings, this work builds upon this idea by extending it to the electrical energy footprint.

The GP-based multi-objective BO scheme was used to determine which set of hyperparameters and neural architecture~\cite{elsken2019neural} 
allows the MLP training phase to achieve both the lowest prediction error on test data and computational energy footprint. Thus, the Adam optimization process corresponds to the black-box evaluation of this parameter optimization framework. 
Table~\ref{table:MnistHPOformulation} gives the objectives to minimize and the considered design variables. The hierarchical GP setup allows for customizing the number of neurons in each layer, while the regular GP forces a fixed number of neurons in all layers. A budget of 200 expensive black-box evaluations was considered, with 10 initial DoE points given by the LHS and 190 BO iterations. Overall, the 200-point HPO process runs for approximately 5 hours. 

\begin{figure}[htbp]
    \hspace{-1cm}
    \includegraphics[width=15cm, height=9cm]{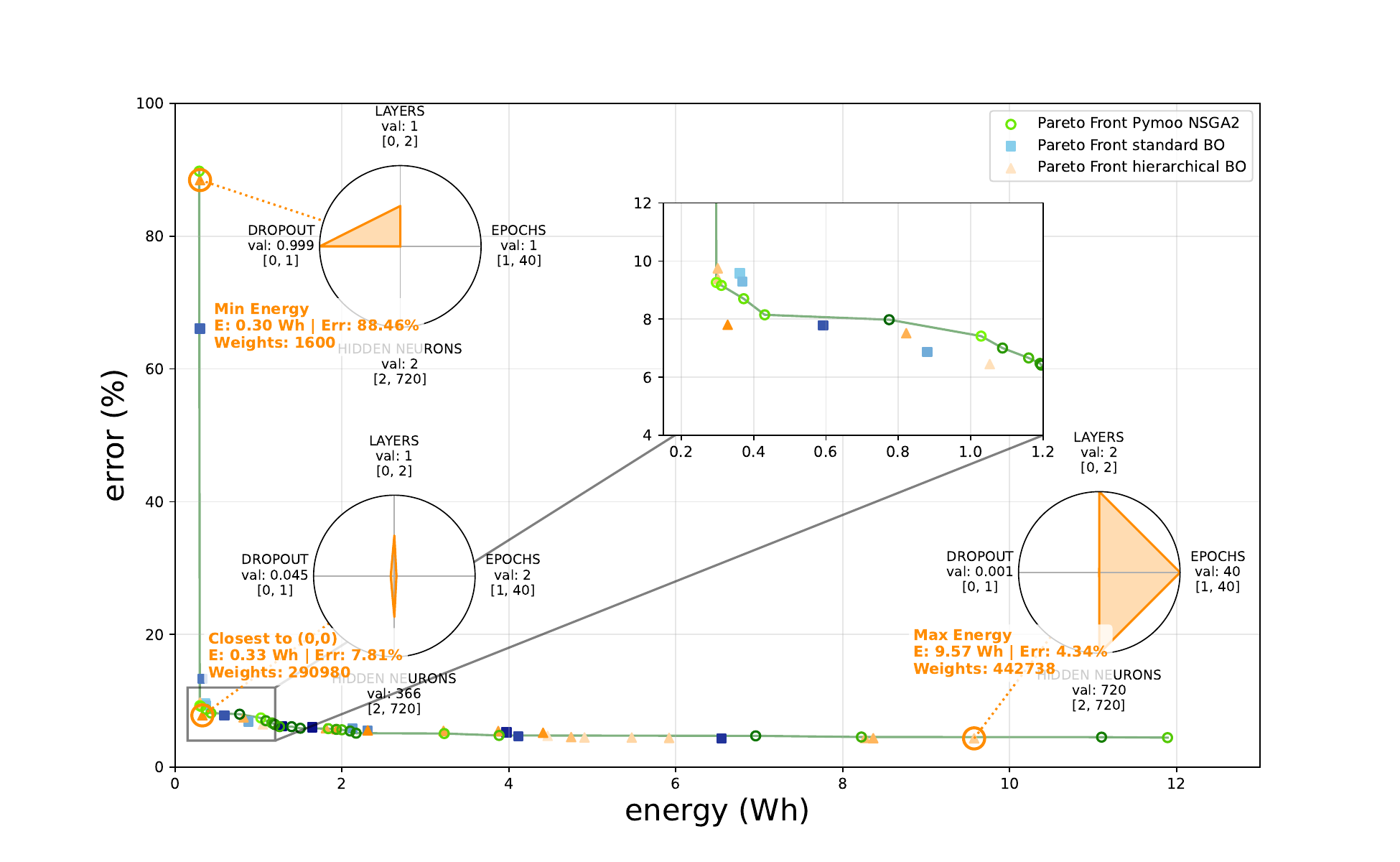}
    \caption{Pareto front between computational energy footprint and error on test data across 1 repetition, with 3 samples of the Pareto set.}
    \label{fig:paretofrontmnist}
\end{figure}

The Pareto fronts for each optimization run are given in Figure~\ref{fig:paretofrontmnist}. The used reference point is 13 Wh and 100 \% error. Three specific hyperparameter configurations from the Pareto set, obtained from the hierarchical BO with EHVI acquisition function run, are also highlighted. Table~\ref{table:mnistresultshpo} compares the convergences between the regular and hierarchical approaches according to whether the EHVI or PI acquisition functions were used during the optimization run. The NSGA-II~\cite{deb2002fast} algorithm, whose implementation comes from Pymoo~\cite{pymoo}, represents the baseline for comparisons. It was given a 2000 black-box budget to ensure good convergence for comparison while staying under 24 hours of computing. The hypervolume metric is used to compare convergences, based on normalized errors and energies. 

Computer experiments successfully resulted in identifying a Pareto front between error on test data and computational energy footprint. As expected, Table~\ref{table:mnistresultshpo} shows that the BO framework correctly converged within a lower budget than NSGA-II. Many parameter configurations have been identified as Pareto optimal, globally stating that the more energy consumed during the training phase, the better the test prediction. Furthermore, configurations close to the origin in Figure~\ref{fig:paretofrontmnist} present low error while being energy-efficient. Up to 96\% computational energy can be saved from increasing error from 4.34\% to 7.81\%, highlighting the existence of interesting tradeoffs to find among parameter setting possibilities. 
Additionally, another more complex and computationally expensive HPO problem is addressed.

\begin{table}[bh!]
    \centering
    \begin{minipage}[t]{0.48\textwidth}
        \caption{Frugal hyperparameter optimization (HPO) problem. Roles account only for the hierarchical formulation.}     
        \label{table:MnistHPOformulation}
    \end{minipage}
    \hfill
    \begin{minipage}[t]{0.48\textwidth}
        \caption{Convergence comparisons, using hypervolume metric (HPV). The best value is highlighted in bold.}
        \label{table:mnistresultshpo}      
    \end{minipage}
    \begin{minipage}[t]{0.48\textwidth}
        \centering
        \resizebox{\linewidth}{!}{%
        \begin{tabular}{llll}
        \hline
        Objective function & Type &  &  \\\hline
        error on test data & cont. &  &  \\
        comput. energy & cont. &   &  \\\hline
        Design variable & Type & Role & Range  \\\hline
        epochs & int. & neutral & [1,40]\\
        dropout & cont. & neutral & (0,1) \\
        \# of layers & int. & meta & [1,5] \\
        \# of neurons / layer & int. & decreed & [2, 512] \\\hline
        \end{tabular}%
        }
    \end{minipage}
    \hfill
    \begin{minipage}[t]{0.48\textwidth}
        \centering
        \resizebox{\linewidth}{!}{%
        \begin{tabular}{llcc}
        \hline
        Optimizer & $\alpha$ & Budget & HPV  \\\hline
        NSGA2 & evol. & 2000 & 0.928 \\
        Regular BO & EHVI & 200 & 0.927 \\
        Regular BO & PI & 200  & 0.928 \\
        Hierarchical BO & EHVI & 200 & 0.929 \\
        Hierarchical BO & PI & 200 & \textbf{0.930} \\\hline
        \end{tabular}%
        }
    \end{minipage}
\end{table}

\subsection{Engineering application: parameter settings for efficient 10-bar truss mass minimization}
\label{truss}
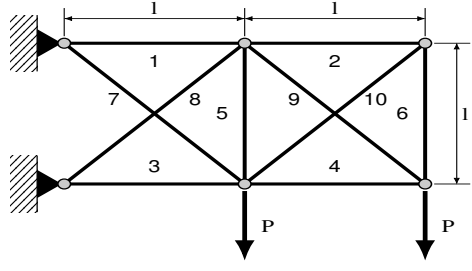
\begin{figure}[htbp]
    \centering
        \resizebox{0.475\linewidth}{3.5cm}{%
        \begin{tikzpicture}[
            joint/.style={circle, draw=black, fill=gray!40, inner sep=0pt, minimum size=8pt, line width=1pt},
            member/.style={line width=2.5pt},
            dimen/.style={<->, >=Latex, thin},
            load/.style={-{Latex[length=6mm, width=4.5mm]}, line width=3pt}
        ]

            \coordinate (A) at (0, 0);   
            \coordinate (B) at (0, 4);   
            \coordinate (C) at (4, 0);   
            \coordinate (D) at (4, 4);   
            \coordinate (E) at (8, 0);   
            \coordinate (F) at (8, 4);   
            
            \fill[pattern=north east lines] (-1.2, -0.8) rectangle (-0.6, 0.8);
            \draw[thick] (-0.6, -0.8) -- (-0.6, 0.8);
            \fill[black] (-0.6, 0.4) -- (-0.05, 0) -- (-0.6, -0.4) -- cycle;
            
            \fill[pattern=north east lines] (-1.2, 3.2) rectangle (-0.6, 4.8);
            \draw[thick] (-0.6, 3.2) -- (-0.6, 4.8);
            \fill[black] (-0.6, 4.4) -- (-0.05, 4) -- (-0.6, 3.6) -- cycle;
            
            \draw[member] (B) -- (D);
            \draw[member] (D) -- (F);
            \draw[member] (A) -- (C);
            \draw[member] (C) -- (E);
            \draw[member] (C) -- (D);
            \draw[member] (E) -- (F);
            \draw[member] (B) -- (C);
            \draw[member] (A) -- (D);
            \draw[member] (D) -- (E);
            \draw[member] (C) -- (F);
            
            \node[joint] at (A) {};
            \node[joint] at (B) {};
            \node[joint] at (C) {};
            \node[joint] at (D) {};
            \node[joint] at (E) {};
            \node[joint] at (F) {};
            
            \draw[load] (4, -0.2) -- (4, -2.2) node[midway, right=2mm] {\Large P};
            \draw[load] (8, -0.2) -- (8, -2.2) node[midway, right=2mm] {\Large P};
            
            \draw (0, 4.2) -- (0, 5.0);
            \draw (4, 4.2) -- (4, 5.0);
            \draw (8, 4.2) -- (8, 5.0);
            \draw[dimen] (0, 4.7) -- (4, 4.7) node[midway, above] {\Large l};
            \draw[dimen] (4, 4.7) -- (8, 4.7) node[midway, above] {\Large l};
            
            \draw (8.2, 0) -- (9.0, 0);
            \draw (8.2, 4) -- (9.0, 4);
            \draw[dimen] (8.7, 0) -- (8.7, 4) node[midway, right] {\Large l};
            
            \node at (2, 3.5) {\Large \textsf{1}};
            \node at (6, 3.5) {\Large \textsf{2}};
            \node at (2, 0.5) {\Large \textsf{3}};
            \node at (6, 0.5) {\Large \textsf{4}};
            \node at (3.5, 2) {\Large \textsf{5}};
            \node at (7.5, 2) {\Large \textsf{6}};
            \node at (1.1, 2.4) {\Large \textsf{7}};
            \node at (2.9, 2.4) {\Large \textsf{8}};
            \node at (5.1, 2.4) {\Large \textsf{9}};
            \node at (6.9, 2.4) {\Large \textsf{10}};
        \end{tikzpicture}
        }
        \caption{10-bar truss structure subjected to two external loads P. Axial stresses are constrained within the material yield stress.}
        \label{fig:trussschema}
\end{figure}

The problem addressed is a structural optimization problem involving a 10-bar truss, illustrated in Figure~\ref{fig:trussschema}, whose mass is minimized subject to the $i$-th maximal static structural constraint with respect to the cross-sectional area of the $i$-th bar. Previous works solved it using deep Gaussian processes (DGP)~\cite{damianou_deep_2013} in a constrained BO framework~\cite{rostum2026comparative}, to capture non-stationary and complex patterns among the structural constraints using a cascade of GP. However, the convergence trajectory depends heavily on the prediction quality of the DGP. Predictive performance fluctuates according to the DGP parameter settings for the training phase, the initialization of hidden layers, and the location of the inducing points. Furthermore, the number of layers and GP nodes, the size of the initial DoE of the cross-sectional areas, and the Monte Carlo sampling parameters significantly impact the quality of the predictions.
Therefore, this second application formulates another HPO problem, described in Table~\ref{table:TrussHPOFormulation}, that minimizes the optimum mass value alongside the computational energy footprint. The truss mass minimization process here corresponds to the black-box evaluation. 

Finally, the HPO algorithm runs 20 iterations using either the EHVI or PI criterion with 10 initial points that were sampled using LHS. Table~\ref{table:resultshpotrussconvergence} compares the hypervolume convergences between BO using the EHVI or PI acquisition function, while Table~\ref{table:resultshpotruss} highlights an example of the gains in objective values obtained using a Pareto optimal configuration. The reference point for the hypervolume evaluations was 3500 Wh and 10000 kg, projected in a normalized space alongside the mass and energy outputs among the Pareto front. In Tables~\ref{table:TrussHPOFormulation} and~\ref{table:resultshpotruss}, the baseline values were drawn from~\cite{rostum2026comparative}, where the epoch value was reduced from 10000 to 1000 for time-consuming reasons.

Numerical results successfully highlighted Pareto optimal energy-effective convergence trajectories. In this case, some hyperparameter configurations even resulted in a better mass minimum with a lower energy computational footprint than the empirically tuned configuration. Indeed, Table~\ref{table:resultshpotruss} shows that the computational energy consumption has been reduced by 91\%, while achieving a 39\% lower feasible mass minimum. Therefore, model selection strongly influences the balance between frugality and performance during optimum searches, and significant enhancements are found using automated methods.    


\begin{table}[htbp]
    \centering
    \begin{minipage}[t]{0.48\textwidth}
        \caption{Frugal HPO problem. Other parameters are equal to values in~\cite{rostum2026comparative}.}
        \label{table:TrussHPOFormulation}
        \centering
        \resizebox{\linewidth}{2.25cm}{%
        \begin{tabular}{llll}
        \hline
        Objective function & Type & &  \\\hline
        feasible mass min. & cont. &  & \\
        comput. energy & cont. &  & \\\hline
        Design variable & Type & Range & Baseline \\\hline
        epochs & int. & [20, 1000] & 1000 \\
        learning rate & cont. &  [0.01, 0.1] & 0.01\\
        multistarts & int. & [3, 15] & 3 \\
        layers & int. & [1,2] & 2 \\
        $n_{DoE}$ & int. & [10, 100] & 100 \\
        $n_{sample} \  { \small \text{hidden layers}}$ & int. & [3,20] & 20 \\
        $n_{sample}\  {\small \text{output layer}}$ & int. & [20,200] & 100 \\\hline
        \end{tabular}%
        }
    \end{minipage}%
    \hfill
    \begin{minipage}[t]{0.48\textwidth}
        
        \caption{Convergence comparisons. Bold text corresponds to the best value.}
        \label{table:resultshpotrussconvergence}
        \centering
        \resizebox{\linewidth}{!}{%
        \begin{tabular}{llll}
        \hline
        Optimizer & $\alpha$ & Budget & Hypervolume  \\\hline
        Regular BO & EHVI & 30 & 0.463 \\
        Regular BO & PI & 30  & \textbf{0.494} \\\hline
        \end{tabular}%
        }
        
        \caption{Example of Pareto optimal outputs, across 5 repetitions. Best value in bold.}
        \label{table:resultshpotruss}
        \centering
        \resizebox{\linewidth}{!}{%
        \begin{tabular}{lllll}
        \hline
        $\alpha$ & E (Wh) & m (kg)  \\
        \hline
        EHVI & \textbf{522.2 $\pm$ 47.5} &   1810 $\pm$ 55.9\\
        PI & 594.9 $\pm$ 222.8 &  \textbf{1742 $\pm$ 10.5}\\
        Baseline & 6459 $\pm$ 2279  & 2858 $\pm$ 500.0 \\
        \hline
        \end{tabular}%
        }
    \end{minipage}
\end{table}
\section{Conclusions and discussions}
\label{discussions}
This study presented a novel computational energy footprint metric to guide optimum searches within a BO framework for engineering design. The approach defined a BO framework that explicitly accounts for the energy consumption of the computations, which was therefore integrated as an additional objective. The primary goal was to prove that the parameter settings of the models matter in ensuring both energy-efficient and accurate convergence.

The results of the benchmark problems demonstrated that the parameter configurations effectively balance the convergence performance with the underlying electrical energy consumption. The results indeed highlighted the existence of a clear Pareto front, revealing tradeoffs between finding accurate optima and the required computational energy. Ultimately, the results demonstrated that the proposed energy consumption metric is representative and helps identify frugal trajectories within the search for optima. 

Despite these promising results, this hyperparameter optimization method carries a high computational cost, as evaluating each parameter configuration currently requires executing the entire optimization problem. In the engineering application, results may not have completely converged even if better configurations than the empirical one were found. To address this issue, future work could rely on faster-to-evaluate, lower-fidelity models to identify optimal hyperparameter configurations more efficiently. Non-myopic acquisition approaches could also be addressed to enable long-term rewards on both performance and energy-efficiency. Furthermore, one could dynamically adapt the models through iterations, updating parameter settings for the best balance between convergence performance and their underlying resource footprint. Because hardware architectures may also influence power consumption, smartly distributing workloads across a cluster of computing nodes could represent another solution to reduce computational overhead. As these infrastructures are integrated within a time-varying electrical grid supply, fluctuations in resource usage efficiency should also be taken into account. Ultimately, beyond viewing frugality simply as energy efficiency, one should also consider the underlying greenhouse gas emissions, water usage, and abiotic resource consumption of computing. 

\section*{Acknowledgements}
The authors are thankful to Heine 
Røstum (Engineering Cybernetics, Norwegian University of Science
and Technology (NTNU), 7034 Trondheim, Norway) for sharing his work, experience, feedback, truss test case and code.

\bibliography{bibliography}
\end{document}